\documentclass[11pt,twocolumn,letterpaper]{article}

\usepackage{cvpr}              

\usepackage{graphicx}
\usepackage{amsmath}
\usepackage{amssymb}
\usepackage{booktabs}
\usepackage{placeins}

\usepackage[pagebackref,breaklinks,colorlinks]{hyperref}
\usepackage{atbegshi}
\usepackage{fancyhdr}
\usepackage{lipsum} 

\usepackage[capitalize]{cleveref}
\crefname{section}{Sec.}{Secs.}
\Crefname{section}{Section}{Sections}
\Crefname{table}{Table}{Tables}
\crefname{table}{Tab.}{Tabs.}

\def\cvprPaperID{*****} 
\def\confName{CVPR}
\def\confYear{2022}

\pdfoutput=1

\begin{document}
\title{Detecting and Localizing Copy-Move and Image-Splicing Forgery}
\author{Aditya Pandey\\ap6624\\New York University\\New York, NY 10003\\
{\tt\small adityapandey@nyu.edu}
\and
Anshuman Mitra\\
am11058\\
New York University\\
New York, NY 10003\\
{\tt\small am11058@nyu.edu}
}
\maketitle

\begin{abstract}
   \textbf{In the world of fake news and deepfakes, there have been an alarmingly large number of cases of images being tampered with and published in newspapers, used in court and posted on social media for defamation purposes.}
   
   \textbf{Detecting these tampered images is an important task and one we try to tackle. 
   In this paper, we focus on the methods to detect if an image has been tampered with using both Deep Learning and Image transformation methods and comparing the performances and robustness of each method. We then attempt to identify the tampered area of the image and predict the corresponding mask.}
   \textbf{Based on the results, suggestions and approaches are provided to achieve a more robust framework to detect and identify the forgeries. }\\
   \textbf{Keywords: Forgery Detection; Image Forgery; Splicing; Copy Move; Deep Learning; Image Forensics;}
\end{abstract} \\
\section{Introduction}
\label{sec:intro}

In today's world, every user has access to some form of easy to use image editing application which has made modifying images extremely easy. While the human eye is capable of identifying obvious changes or modifications to an image, it is not difficult to create virtually undetectable manipulations using software such as Adobe Photoshop, MS Paint, Corel Draw, etc. \cite{Photoshop}.
There are many different types of image forgery/manipulation such as copy-move forgery, image splicing, cropping, image masking, blurring, etc. For the purposes of this paper, we would be focusing on the latter two types of image manipulation - image splicing and copy-move forgery. 

Image splicing is the process of modifying an image by pasting a cropped part of another image and pasting it into the first image. This is a very common form of forgery and is often followed by some form of image augmentation such as blurring or smoothing to hide the manipulation.
Copy move forgery on the other hand is when an image is tampered by copying and pasting a small region from an image into the same image.
\begin{center}
    \includegraphics[scale=0.4, width=80mm]{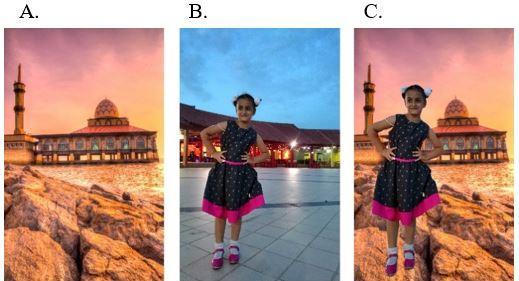}\\
    An example of image splicing \cite{example1} 
\end{center}
\begin{center}
    \includegraphics[scale=0.2]{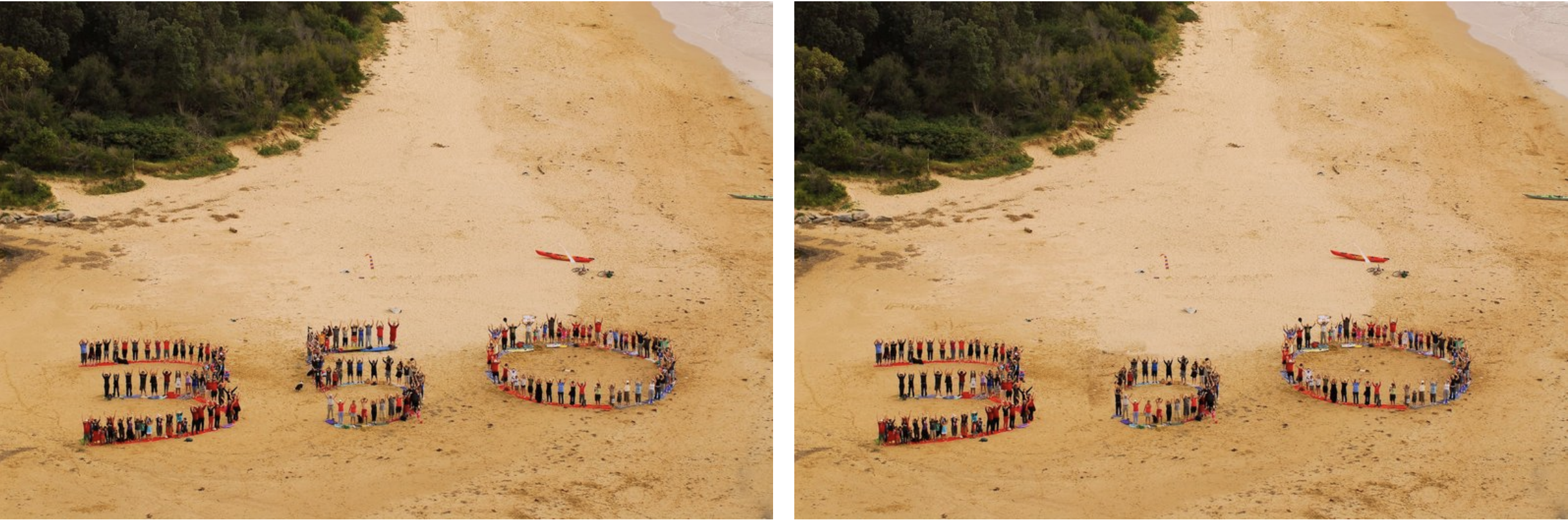}\\
    An example of copy-move forgery \cite{Copy-Move}. \\
    Note how the 5 has been modified in a way that is virtually indistinguishable to the eye.
\end{center}
\newpage
As mentioned in the abstract, there are essentially two parts to identifying a forged image - the first one is to classify an input image as forged/tampered or not. This is a classification task and is what is called forgery detection.
The second part of the problem statement is to identify the 'forged' part of the image i.e. to predict the mask of the forged image which is a black and white image (output) which describes the area of the image that has been 'tampered' with.
\begin{center}
    \includegraphics[scale=0.225]{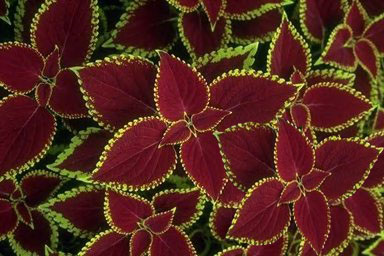}
    \includegraphics[scale=0.3]{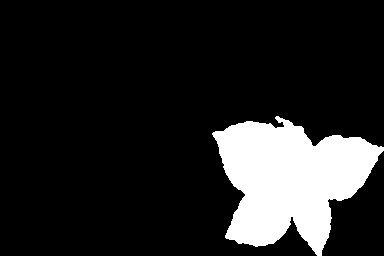}\\
    A tampered image alongside the mask indicating the forged area.
\end{center}

\section{Related Work}
\label{sec:Related Work}
Most of the work done in the image tampering detection space has used image transformations as the backbone. Images are essentially signals and identifying patterns of contours, textures, curves, etc. give us a robust enough feature set to classify images as tampered or authentic. 
The work done by Mohammad et. al.\cite{Mohammad}, Shilpa et al.\cite{shilpa} and Alahmadi et. al.\cite{alahmadi} use discrete cosine transform (DCTs) to extract features to detect forgery. The application of DCT here is to essentially convert an image into a feature set made of the sinusoids of varying magnitudes and frequencies. A slightly different approach using Discrete Wavelet Transforms (DWTs) with Local Binary Patterns was attempted by Mandeep et. al. \cite{Mandeep}.\\
Rao and Ni \cite{rao} presented a convolutional neural network (CNN) based method to detect forgeries by replacing the first layer weights of the CNN with high pass filters to capture the features from input images. This feature set is then fed into an SVM classifier to perform the classification task. On the deep learning front, Jaiswal et. al. \cite{jaiswal} proposed a Deep learning model based on ResNet-50 for feature extraction from the CASIA2.0 dataset to extract features from the input images and feed into different binary classifiers. This method is devoid of any sort of image transformation or filtering and is a pure Deep Learning approach to identifying tampered images. 
The work of Sudiatmika et. al. \cite{Sudiatmika} uses a VGG16 based model which is trained on input images after Error Level Analysis (ELA) has been performed 
\\

To identify the areas of the image that are tampered with, most approaches use a deep neural network to perform the task. 

Davide et. al. \cite{davide} use noiseprint, a recently proposed CNN-based
camera model fingerprint to localize image forgeries.
Yao, Hongwei et al. \cite{yao} use a reliability fusion map to localize the forgery using fingerprints acquired by images from different camera models with a CNN. 
Sri Kalyan et. al. \cite{SriKalyan} used a generative adversarial network (GAN) based model to learn features of authentic images followed by SVMs to learn distribution and detect forged images as anomalies.
Convolutional Neural Networks were used by Liu et. al. \cite{liu} along with segmentation-based analysis to locate tampered areas in images.

Photo Response Non-Uniformity (PRNU) analysis is one of the most popular approaches to perform tampering localization. Paweł et. al \cite{pawel} take a 
multi-scale fusion approach by combining multiple candidate tamper maps using sliding windows of varying sizes and extend it by introducing neighbour interactions between maps.
Chakraborty et. al. \cite{Chakraborty} also use PRNU analysis and treat the localization problem as a probabilistic binary labeling task.

The UNet architecture created by Ronneberger et. al. \cite{Ronneberger}, initially developed for biomedical image segmentation forms the basis of a number of recent forgery localization solutions.
UNets are popular due to the lack of camera point information which prevents the application of PRNU based approaches.
Xiuli et. al. \cite{Xiuli} attempt to combine UNets with encoders to localize Image Splicing Forgery. 
Ding, Hongwei et. al. \cite{Hongwei} proposed a method based on dual-channel U-Net aka DCU-Net which consists of three parts: encoder, feature fusion, and decoder.
\section{Approach}
\label{sec:approach}
We approach our problem statement in 2 distinct parts - first being detecting forgeries i.e. a binary classification task to predict whether an input image has been tampered with or not. The second part is to predict the mask of an input image. Here, we are assuming that each input image, whether it is authentic or tampered will have a mask present.

\subsection{Detection of Tampered Image}
The first approach we take to detect whether an image has been tampered is by using the properties of JPEG compression. We perform Error Level Analysis (ELA) which identifies areas in the image that are at different compression levels.
ELA is performed by compressing an image in JPEG format and finding the differences between the original image and the compressed image.
When performing ELA, areas of the image that are smooth and uniform like sky, walls, etc will have a lower ELA result.
\begin{center}
\includegraphics[scale=0.4]{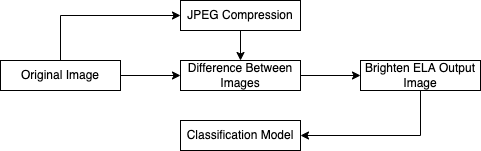}
\end{center}
Using Error Level Analysis identifies parts of the image which were forged as those areas have a higher ELA values. Using this as our feature set, we train our classifier models - we use a Support Vector Classifier.
Once we have our featureset, we also use different types of Neural Networks. We start off with a simple Conv2D-block based CNN with 4 Convolutional blocks. 
We also try and use ResNet architecure using Transfer Learning to form a Deep Network.
Finally, we experiment with an LSTM model to compare the results of our approaches. We also test our dataset by performing Cosine Transformations with Local Binary Patterns similar to \cite{Mandeep} using a SVM and a simple neural network (non-CNN). 
\begin{center}
\includegraphics[scale=0.4]{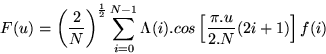}
\end{center}
To achieve this, we use the DCT formula (expanded for a 2D block which is created by performing LBP on multiple sub blocks of the original image)
\begin{center}
\includegraphics[scale=0.4]{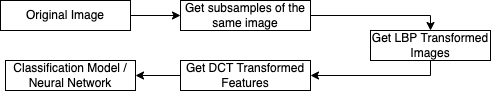}
\end{center}
To test the effect of different image effects, we also compare the performance of the models on the blurred version of the dataset. Blurring an image removes some of the noise and distinct features and does indeed affect the performance of our models.

\subsection{Identifying Tamper Mask}
We also attempt to identify the mask of a tampered image. A mask is essentially a binary image which is capable of hiding (or selecting) parts of an image to apply transformations to. In our problem statement, the mask refers to the part of the image that has been tampered with. 

To solve this, we base our experiments on the UNet Architecture proposed by Ronneberger et. al. \cite{Ronneberger}. We first convert our inputs to a feature vector using Error Level Analysis before feeding it into our model. 
\begin{center}
\includegraphics[scale=0.4]{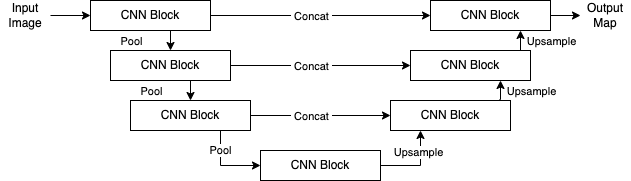}
Basic Structure of a UNet Architecture
\end{center}
We experiment with two approaches for our use case - one with using only the tampered images and it's corresponding mask \cite{casia2_groundtruth} and the other in which we incorporate a mask for each authentic image i.e. we create an empty mask for an authentic image and train our model using the combined dataset.
For the internal architecture of the UNet, we compare the results of using Conv2D blocks to build our model with a model based on the ResNet-101 architecture proposed by Kaiming et. al. \cite{Kaiming}.
Finally we reconstruct our masks and analyze the results by visual inspection as well as the corresponding loss values of the models trained on both versions of the dataset.

\section{Experiments and Results}
\label{sec:experiments}

\subsection{Training and Evaluation}
We trained our models on a combination of NYU Greene Cluster and Google Colab. The Colab instances use the Tesla P100 GPU for all our experiments an 25GB of RAM. To try and replicate the memory bottleneck (if any), our Greene Cluster uses an NVIDIA V100 GPU with 32GB of RAM. We used Keras to create our model with a TensorFlow backend.

For the Forgery Detection, we use the Accuracy, F-Score for each class (Tampered/ Not Tampered) and Weighted average to compare the results and performances of each model. Being a binary classification task, we perform quantitative comparision between different approaches.

For the identification of the mask of a tampered image, a combination of loss as well as qualitative comparision is used to compare the approaches. Each model minimizes the validation loss (where the validation data is 0.2 of the dataset).
We resize all images and masks to dimensions 128X128 which is the input size for our UNet. 

\subsection{Dataset}
For the purposes of this paper, we will use the CASIA2.0 dataset \cite{casia2} introduced by Jing Dong et al. with its corresponding ground truth masks \cite{casia2_groundtruth} by Nam Thanh et al. 
The dataset contains 7408 authentic images and 5123 tampered images out of which 3295 are of copy-move type and 1828 are spliced images. 
Each tampered image has a corresponding binary mask indicating the area of tamper. 
For training purposes, we use the created masks for authentic images (which is just an empty 2D array of 0s). Further preprocessing steps include testing the model with black and white images, by introducing ablations like blurred images and generating flipped and rotated samples to increase the training size.
For the forgery detection part, we also scale the dataset in the process of performing Error Level Analysis.
The output of the DCT transform is also reduced to 0-1 scale to perform effective SVC and NNet classification. 

\begin{table*}[!h]
\begin{center}
\caption{Detecting Forged Images}
\begin{tabular}{||c| c c c c||} 
 \hline\hline
 Model & Accuracy & F-Score 0 & F-Score 1 &  Weighted Score\\ [0.5ex] 
 \hline\hline
 Digital Cosine Transform  & 0.92 & 0.93 & 0.88 & 0.91 \\ 
 with SVC &  &  & &  \\ 
 \hline
 \textbf{Digital Cosine Transform}  & \textbf{0.93} & \textbf{0.94} & \textbf{0.91} & \textbf{0.93} \\ 
 \textbf{with Simple NNet} &  &  & &  \\ 
 \hline
 ELA + Basic CNN & 0.88 & 0.92 & 0.76 & 0.86 \\
 \hline
 ELA + ResNet CNN & 0.89 & 0.92 & 0.80 & 0.88 \\
 \hline
 ELA + Basic CNN & 0.91 & \textbf{0.94} & 0.83 & 0.90 \\
 with BatchNorm &  &  &  &  \\ 
 \hline
 \hline
\end{tabular}
\end{center}
\end{table*}
\subsection{Forgery Detection}
As we can see from the results in Table 1, the Digital Cosine Transform Feature Extraction combined with a simple neural network model gives us the best results (in terms of accuracy as well as the Recall and Precision for individual classes)

Where the results get interesting is detecting forgeries in Images that have been processed further after the initial tamper.
\textbf{Blurring} an image blends the tampered bit into the original image. As we are dealing with Copy-Move and Image-Splicing forgery, this blur would affect the frequency and general properties of the image as the blurring operation sets each pixel as the average of surrounding pixels.

As we can see from the results table, The Digital Cosine Transform comes up as the best result out of all our approaches. Combined with a Basic Neural network, we are able to achieve a robust F-Score for each class. Error Level Analysis also gives us comparable results when used with a CNN which has batch norm but favours the class 0.
The story is even more distinct when working with input images that have been augmented with shearing or blurred. The DCT based models are still the best performing models although there is a loss in accuracy here. The models based on CNNs perform much worse here - all but failing to identify the tampered images which have been bblurred (identifying the shear transform images to some extent).

\subsection{Localization}
While attempting to create a localization model, we use two approaches based on UNETs. The first one takes only the forged images as input and tried to predict the mask based on the calculated input feature map. 
As we can see from the results, while there is some mask identified by the model, on further visual inspection the performance of this model is not upto par.
\begin{center}
    Approach 1: The original pask followed by the predicted mask (output+binary output) 
    \includegraphics[scale=0.75]{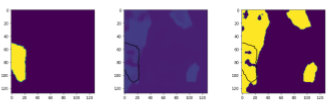}
\end{center}
The next approach involved passing two input images i.e. the original image along with the ELA-transformed image. The output/target here is still the corresponding mask of the image.
The model here also uses a ResNet101 backend as the base model.
As we can see from the examples as well as the outputs of input test images, there is a clear improvement in the mask prediction as compared to the first attempt.
\begin{center}
    Approach 2: The original pask followed by the predicted mask (output+binary output) 
    \includegraphics[scale=0.75]{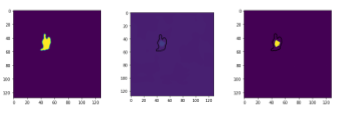}
\end{center}

\begin{center}
    \includegraphics[scale=1]{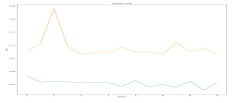}\\
    Loss Curve for CNN Based UNet
    \includegraphics[scale=1]{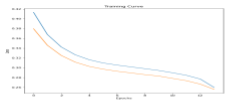}\\
    Loss Curve for ResNet Based UNet \\
    (Orange is the validation loss and blue is the train loss)
\end{center}

\begin{center}
\captionof{table}{Detecting Forged Images with Augmentation, Ablations} 
\begin{tabular}{||c| c c ||} 
 \hline\hline
 Model & CNN UNet & ResNet UNet\\ [0.5ex] 
 \hline\hline
 Validation Loss & 0.156 & 0.26 \\
  \hline
 Training Loss & 0.162 & 0.21 \\
  \hline
 Time Per Epoch & 93s & 161s \\
  \hline
 Convergence Epoch & 17 & $>$25 \\
  \hline
  \hline
\end{tabular}
\end{center}

\begin{table*}

\begin{center}
\caption{Detecting Forged Images with Augmentation/Ablations}
\begin{tabular}{||c| c c c c||} 
 \hline\hline
 Model & Accuracy & F-Score 0 & F-Score 1 &  Weighted Score\\ [0.5ex] 
 \hline\hline
 ELA + Basic CNN &  &  &  &  \\ 
 + BatchNorm + Augmentation & 0.62 & 0.71 & 0.45 & 0.65 \\
 (Shear and Rotation) &  &  &  &  \\ 
 \hline
 ELA + Basic CNN & 0.80 & 0.89 & 0.11 & 0.72 \\
 + Blur &  &  &  &  \\ 
 \hline
 ELA + Basic CNN & 0.77 & 0.88 & 0.06 & 0.69 \\
 + Reduce Channels &  &  &  &  \\ 
 \hline
Digital Cosine Transform  & 0.87 & 0.89 & 0.86 & 0.87 \\ 
 with SVC &  &  & &  \\ 
 \hline
 \textbf{Digital Cosine Transform}  & \textbf{0.90} & \textbf{0.90} & \textbf{0.88} & \textbf{0.90} \\ 
 \textbf{with Simple NNet} &  &  & &  \\ 
\textbf{ + Blur} &  &  & &  \\ 
 \hline
  \hline
\end{tabular}
\end{center}

\end{table*}

\section{Conclusion}
This work demonstrates the effectiveness of using Deep Learning/Machine learning models along with image transformations in identifying forgery in images. As seen from the results, the Digital Cosine Transformation is an extremely powerful tool in identifying images where there is an anomaly in the frequency in the image.

We can see that introducing filters and some forms of augmentations make it more difficult to identify tampering in images and aligns with the fact that blurring and modifying tampered images which remain an effective way of hiding forgeries.

On the front of identifying areas of the image that have been tampered, UNET based models are effective in identifying pars of the tamper and output masks. Performing some form of analysis on the raw image (ELA or DCT) remains an effective step in making the model more robust.

\section{Future Work}
We can see that our DCT model while being relatively robust to blurring, other ones do perform poorly when the number of added embellishments are high. Performing further signal pre-processing on the images (such as Digital Wave Transform, Fourier Transform, etc.) might result in an even more robust model.
The work can also be expanded to include forgeries outside of the copy-move and image-splice types.

On the localization front, increasing our data size by introducing more tampered images and their masks could increase the performance of the model. We should also study the affect of introducing more information based on the frequency and distribution of images into out UNet model which might increase the performance of the same (this could be an interesting approach looking at the performance of using Transforms in the detection task).

{\small
\bibliographystyle{ieee_fullname}
\bibliography{egbib}
}

\end{document}